\documentclass{article} 
\usepackage{iclr2016_conference,times}
\usepackage{xcolor}
\usepackage[colorlinks, linkcolor=black, urlcolor=black, citecolor=black]{hyperref}
\usepackage{caption}
\usepackage{url}
\usepackage{booktabs}
\usepackage{textcomp}
\usepackage{graphicx}
\usepackage{subcaption}
\usepackage{amsmath,amssymb}

\title{Order Matters: Sequence to sequence for sets}

\author{
Oriol Vinyals, Samy Bengio, Manjunath Kudlur\\
Google Brain\\
\texttt{\{vinyals, bengio, keveman\}@google.com} \\
}

\iclrfinalcopy 

\begin{document}

\maketitle

\begin{abstract}
Sequences have become first class citizens in supervised learning thanks to the resurgence of recurrent neural networks. Many complex tasks that require mapping from or to a sequence of observations can now be formulated with the sequence-to-sequence (seq2seq) framework which employs the chain rule to efficiently represent the joint probability of sequences. In many cases, however, variable sized inputs and/or outputs might not be naturally expressed as sequences. For instance, it is not clear how to input a set of numbers into a model where the task is to sort them; similarly, we do not know how to organize outputs when they correspond to random variables and the task is to model their unknown joint probability. In this paper, we first show using various examples that the order in which we organize input and/or output data matters significantly when learning an underlying model.
We then discuss an extension of the seq2seq framework that goes beyond sequences and handles input sets in a principled way. In addition, we propose a loss which, by searching over possible orders during training, deals with the lack of structure of output sets.
We show empirical evidence of our claims regarding ordering, and on the modifications to the seq2seq framework on benchmark language modeling and parsing tasks, as well as two artificial tasks -- sorting numbers and estimating the joint probability of unknown graphical models.
\end{abstract}

\section{Introduction}

Deep architectures have shown in the last few years that they often yield
state-of-the-art performance on several tasks, ranging from image
classification~\citep{batchnorm} to speech recognition~\citep{dnn-speech}.
More recently, recurrent neural networks (RNNs) and variants such as the Long Short Term Memory network (LSTMs) proposed by~\cite{hochreiter1997long}
have shown similar
impressive performance on several inherently sequential tasks. Such examples range
from machine translation~\citep{sutskever2014sequence,bahdanau2014neural},
to image captioning~\citep{vinyalscaptions,baidu_captioning,berkeley_captioning},
speech recognition~\citep{las,bahdanau_speech},
constituency parsing~\citep{grammar} and learning to compute~\citep{execute,pointernets}.
These approaches all follow a simple
architecture, dubbed {\em sequence-to-sequence} (seq2seq), where the input 
is read completely using an encoder, which is either an LSTM when the
input is a sequence, or a convolutional network for images.
The final state of the encoder is then fed to a decoder LSTM whose purpose
is to produce the target sequence, one token at a time.

When the data is naturally organized as a sequence, the sequence-to-sequence
framework is well suited. For example, the chain rule is used to decompose the joint
probability of sequences of words, and can be implemented by an LSTM without making
any conditional independence assumption.
But how should we represent data, either inputs or outputs, for problems where an obvious
order cannot be determined?
For instance, how should we encode a set of numbers when
the task is to sort them? Alternatively, how should we output a set of 
detected objects in an image when there is no specific known order among them?
Does the {\em a priori} choice of ordering of the data to be presented to
the model matter?

The purpose of this paper is two-fold. First, we show that even when
no natural order is known among input or output objects, there might still
be one that yields better performance, hence, {\em order matters}.
Second, we propose two approaches to consider sets either as inputs and/or
outputs in our models and evaluate how they perform on various artificial
and real datasets.

\section{Related Work}

Since sequence-to-sequence models were proposed for machine translation~\citep{sutskever2014sequence,choetal,nal_emnlp}, the research community has proposed several applications in which these models can perform mappings from and/or to sequences. For example, image captioning maps from an image to a sentence~\citep{vinyalscaptions,baidu_captioning,berkeley_captioning}, parsing maps from a sentence to a (linearized) parse tree~\citep{grammar}, and models for computation map from problem statements (e.g. a python program or a set of points on the plane) to their solutions (the answer to the program~\citep{execute}, or the traveling salesman problem tour for the set of points~\citep{pointernets}). It is out of the scope of this paper to review all successful applications of seq2seq, but the list above already includes some non-trivial examples of mapping to/from objects that are not necessarily sequences.

More recently, many related models and key contributions have been proposed, that utilize the concept of external memories, including RNNSearch~\citep{bahdanau2014neural}, Memory Networks~\citep{memory_networks} and Neural Turing Machines~\citep{neural_turing_machines}.
The key element that these models utilize is a reading (or attention) mechanism to read these external memories in a fully differentiable way (though there has also been work with discrete reading mechanism, most notably RL-NTM~\citep{ilyarlntm}).

Unlike traditional structured prediction algorithms~\citep{structured_data}, our approach relies on the chain rule to serialize output random variables through the strong capabilities of LSTM networks to model long-term correlation. Similarly, we do not want to assume a known structured input, as is done for instance with recursive neural networks~\citep{socher} which encode sentences recursively as (given) trees.

\section{Neural Networks for Sequences and Sets}

Let us consider a generic supervised task with a given training set of $n$
pairs $(X^i,Y^i)_{i=1}^n$ where $(X^i, Y^i)$ is the $i^{th}$ pair of
an input and its corresponding target.
The {\em sequence-to-sequence} paradigm corresponds
to tasks where both $X^i$ and $Y^i$ are represented by sequences, of possibly different lengths: $X^i = \{x^i_1,x^i_2,\ldots,x^i_{s_i}\}$ and
$Y^i = \{y^i_1,y^i_2,\ldots,y^i_{t_i}\}$.
In this case, it is reasonable to model each example using the conditional
probability $P(Y|X)$ and to use the chain rule to decompose it as
follows (we drop the example index $i$ in the rest of this section for readability):
\begin{equation*}
P(Y|X) = \prod_{t=1}^{T} P(y_t | y_1, y_2, \ldots, y_{t-1}, X)
\end{equation*}
and implement it as an encoder recurrent neural network (RNN) to read
sequentially each $x_s \in X$ as follows:
\begin{equation}
\label{eq:encoder}
h_s = f_{enc}(h_{s-1}, x_s)
\end{equation}
where $h_s$ is the state of the encoder at time $s$, followed by a decoder RNN to produce each
$y_t \in Y$ one at a time, given the current state $g_t$ and the previous $y_{t-1}$
symbol:
\begin{eqnarray*}
g_1 & = & h_s \\
g_t & = & f_{dec}(g_{t-1}, y_{t-1})
\end{eqnarray*}
\begin{equation}
\label{eq:decoder}
P(y_t| y_1, y_2, \ldots, y_{t-1}, X) = \mbox{softmax}(\mbox{affine}(g_t))\;.
\end{equation}

The use of the chain rule makes this approach assumption free,
so when the input $X$ corresponds to a sequence (like a sentence), it
is reasonable to read it sequentially into an RNN, as in eq.~(\ref{eq:encoder}). However, how should we encode
$X$ if it does not correspond naturally to a sequence? For instance, what
if it corresponds to an unordered set of elements?

Similarly, when the target $Y$ corresponds to a sequence, it is
reasonable to produce it sequentially with an RNN, as in eq.~(\ref{eq:decoder}),
but how should we
produce $Y$ if it does not correspond naturally to a sequence?

Note that sequences can be encoded as sets.
Indeed, if we associate to each element of a sequence the index it occupies in it, forming a tuple, we effectively convert this sequence to a set. For example, the sequence ``I like cats'' becomes the set \{(I,1), (like,2), (cats,3)\} (note that we can permute elements in the set but still recover the original sequence). Although this may be unnecessary in some cases, we argue that, even for sequences, inputting and/or outputting them in a different order could be beneficial. For example, in sorting we may want to employ a divide-and-conquer strategy which finds the median element first (i.e., we may output the solution in neither increasing nor decreasing sequential order).

In the following two sections we discuss how to extend seq2seq to handle input
sets (Section~\ref{sec:input_sets}) and output sets (Section~\ref{sec:output_sets}). We also show the importance of ordering in a variety of tasks in which seq2seq has successfully been applied, and include experimental results to support our claims and extensions to the existing models.

\section{Input Sets}
\label{sec:input_sets}

We first study extensions to encoding (reading) sets. As we discussed in the previous section, sequences can be read with a recurrent neural network which can compress its contents into a vector. An important invariance property that must be satisfied when the input is a set (i.e., the order does not matter) is that swapping two elements $x_i$ and $x_j$ in the set $X$ should not alter its encoding.

A simple approach which satisfies this, and which in fact has been commonly used for encoding sentences, is the bag-of-words approach. In this case, the representation is simply a reduction (e.g., addition) of counts, word embeddings, or similar embedding functions, and is naturally permutation invariant. For language and other domains which are naturally sequential, this is replaced with more complex encoders such as recurrent neural networks that take order into account and model higher order statistics of the data.

An unsatisfying property of using a reduction operation (such as addition) is that it makes the representation quite inefficient: the model operates over a fixed dimensional embedding regardless of the length of the set. It is unlikely that such representation will succeed, as the amount of memory required to encode a length $T$ set (or sequence, for that matter) should increase as a function of $T$. Thus, we argue that even deep convolutional architectures will suffer from this limitation -- though some modifications exist~\citep{maas:2012}.

In our work, we largely rely on attention mechanisms to integrate information from a variable length structure, which we describe in Section~\ref{sec:att}.

\subsection{Input Order Matters}

In this section, we highlight prior work where we observed that the order of inputs impacted
the performance of seq2seq models taking sequences as input.
In principle, order should not
matter when using a complex encoder such as a recurrent neural network, as these are universal approximators that can encode complex features from the input sequence (e.g., n-grams of any order).
We believe that the reason order seems to matter is due to the underlying non-convex optimization and more suitable prior.

The first example which we experimented with was altering the order of sequences in the context of machine translation. In machine translation, the mapping function encodes a sentence in a source language (e.g., English), and decodes it to its translation in a target language (e.g., French). By reversing the order of the input English sentence, \cite{sutskever2014sequence} got a 5.0 BLEU score improvement which allowed them to close the gap between their model -- a fully end-to-end model for machine translation -- and state-of-the-art models which were highly engineered. Similarly, for constituency parsing, in which the mapping is from an English sentence to a flattened version of its constituency parse tree, a 0.5\% absolute increase in F1 score was observed when reversing the English sentence~\citep{grammar}.

Furthermore, if we preprocess the data for, e.g., convex hull computation that was presented in \cite{pointernets} by sorting the points by angle, the task becomes simpler (from $O(n\log n)$ to $O(n)$), and as a result the models obtained are much faster to train and better (increasing accuracy by up to 10\% absolute in the most challenging cases).

All these empirical findings point to the same story: often for optimization purposes, the order
in which input data is shown to the model has an impact on the learning performance.

Note that we can define an ordering which is independent of the input sequence or set $X$ (e.g.,
always reversing the words in a translation task), but also an ordering which is input dependent (e.g., sorting the input points in the convex hull case). This distinction also applies in the discussion about output sequences and sets in Section~\ref{sec:output_order}.

Recent approaches which pushed the seq2seq paradigm further by adding memory and computation to these models allowed us to define a model which makes no assumptions about input ordering, whilst preserving the right properties which we just discussed: a memory that increases with the size of the set, and which is order invariant. In the next sections, we explain such a modification, which could also be seen as a special case of a Memory Network~\citep{memory_networks} or Neural Turing Machine~\citep{neural_turing_machines} -- with a computation flow as depicted in Figure~\ref{fig:read_attend_write}.

\subsection{Attention Mechanisms}\label{sec:att}

Neural models with memories coupled to differentiable addressing mechanism have been successfully applied to handwriting generation and recognition~\citep{graves_thesis}, machine translation~\citep{bahdanau2014neural}, and more general computation machines~\citep{neural_turing_machines,memory_networks}. Since we are interested in associative memories we employed a ``content'' based attention.
This has the property that the vector retrieved from our memory would not change if we randomly shuffled the memory. This is crucial for proper treatment of the input set $X$ as such. In particular, our process block based on an attention mechanism uses the following:

\begin{minipage}{.4\linewidth}
\begin{eqnarray}
q_t &=& LSTM(q^*_{t-1}) \label{eq:first}\\
e_{i,t} & = & f(m_i, q_t) \\
a_{i,t} & = & \frac{\exp(e_{i,t})}{\sum_j \exp(e_{j,t})} \\
r_t & = & \sum_i a_{i,t} m_i \label{eq:blend} \\
q^*_t &=& [q_t\ r_t] \label{eq:last}
\end{eqnarray}
\end{minipage}%
\hspace{.1\linewidth}
\begin{minipage}{.5\linewidth}
\centering
\includegraphics[width=\linewidth]{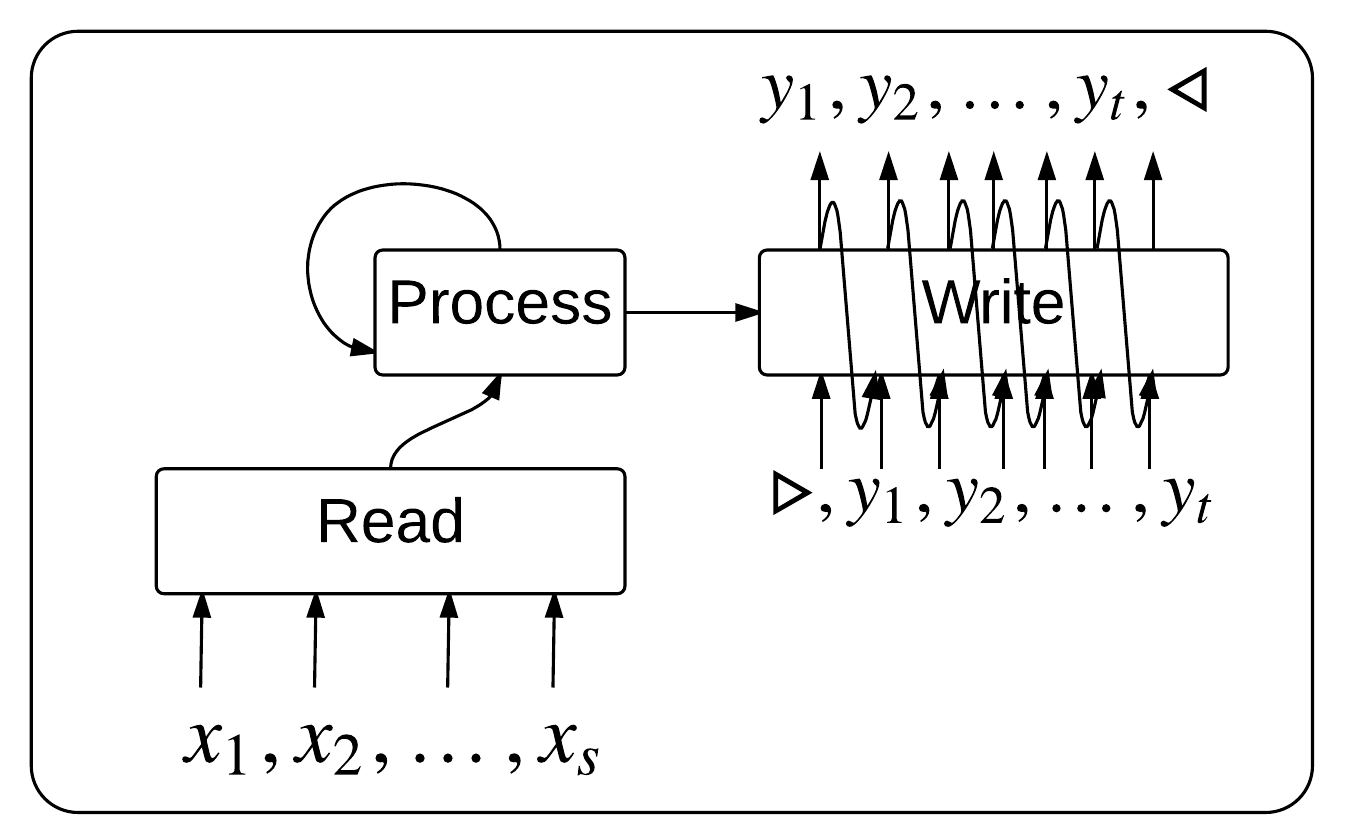}
\captionof{figure}{\label{fig:read_attend_write}The Read-Process-and-Write model.}
\end{minipage}

where $i$ indexes through each memory vector $m_i$ (typically equal to the cardinality of $X$), $q_t$ is a query vector which allows us to read $r_t$ from the  memories, $f$ is a function that computes a single scalar from $m_i$ and $q_t$ (e.g., a dot product), and $LSTM$ is an LSTM which computes a recurrent state but which takes no inputs. $q^*_t$ is the state which this LSTM evolves, and is formed by concatenating the query $q_t$ with the resulting attention readout $r_t$. $t$ is the index which indicates how many ``processing steps'' are being carried to compute the state to be fed to the decoder. Note that permuting $m_i$ and $m_{i'}$ has no effect on the read vector $r_t$.

\subsection{Read, Process, Write}

Our model, which naturally handles input sets, has three components (the exact equations and implementation will be released in an appendix prior to publication): 

\begin{itemize}
\item A \emph{reading} block, which simply embeds each element $x_i \in X$ using a small neural network onto a memory vector $m_i$ (the same neural network is used for all $i$).
\item A \emph{process} block, which is an LSTM without inputs or outputs performing $T$ steps of computation over the memories $m_i$. This LSTM keeps updating its state by reading $m_i$ repeatedly using the attention mechanism described in the previous section. At the end of this block, its hidden state $q^*_T$ is an embedding which is permutation invariant to the inputs. See eqs.~(\ref{eq:first})-(\ref{eq:last}) for more details.
\item A \emph{write} block, which is an LSTM pointer network~\citep{pointernets} that takes in $q^*_T$ (as the context it needs from which to produce the output from the input set), and points at elements of $m_i$ (implicitly, $x_i$), one step at a time. The original work in \cite{pointernets} used a pointer mechanism which, instead of issuing a readout of memory by a weighted sum with a soft pointer (see eq.~\ref{eq:blend}), it uses the pointer as part of the loss. We extended this by adding an extra attention step before the pointer (we called this glimpse). This is related to the process block described above, but with the difference that the attention reads happen interleaved between each pointer output. As described later in the results, we found these two mechanisms to complement each other.
\end{itemize}

The architecture is depicted in Figure~\ref{fig:read_attend_write} and can be seen as a special case of a Neural Turing Machine or Memory Network. It satisfies the key property of being invariant to the order of the elements in $X$, thus effectively processing the inputs as a set. Also note that the \emph{write} component could simply be an LSTM if the outputs were from a fixed dictionary. For this model, though, we study combinatorial problems where the outputs are pointers to the inputs, so we use a pointer network.

\subsection{Sorting Experiment}\label{sec:sorting}

In order to verify if our model handles sets more efficiently than the vanilla seq2seq approach,
we ran the following experiment on artificial data for the task of sorting
numbers: given $N$ unordered random floating point numbers between 0 and 1, we return them in a
sorted order. Note that this problem is an instance of set2seq. We used the architecture defined in
Figure~\ref{fig:read_attend_write}, where the {\em Read} module is a small multilayer perceptron
for each number, the {\em Process} module is an attention
mechanism over the read numbers, implemented as $T$ steps over an LSTM
with no input nor output, but attending the input embeddings, followed
by an LSTM to produce indices in the input numbers with a pointer network \citep{pointernets}, in the proper sorted order.
We also compared this architecture with a vanilla seq2seq
architecture made of an input LSTM connected to an output LSTM which produces
indices in the input numbers with a pointer network (Ptr-Net). Note that the only difference
between these two models is the encoding of the set using either an LSTM (as in previous work),
or with the architecture proposed in the previous section. We ran
multiple experiments, varying the number $N$ of numbers to sort, as well
as the number $T$ of processing steps of the Read, Process, Write model.

The out-of-sample accuracies (whether we succeeded in sorting all numbers or not)
of these experiments are summarized in
Table~\ref{tab:sorting}. We can see that the baseline pointer network LSTM input model
is better than the Read-Process-and-Write model when no processing steps ($P=0$ step)
are used, but as soon as at least one processing step is allowed, the
performance of the Read-Process-and-Write model gets better, increasing with
the number of processing steps. We can also see that, as the size of the
task (expressed in the number of elements to sort $N$) grows, the performance
gets worse, as expected. Also note that with 0 processing steps and 0 glimpses, the writing module is
effectively unconditioned on $X$ and has to ``blindly'' point at the elements of $X$. Thus,
it is unsurprising to see it performing worse than any other model considered in Table~\ref{tab:sorting}.
Lastly, equipping the writing module with glimpses (i.e., adding an attention mechanism prior to ``pointing'') improves both the baseline model (Ptr-Net), and our proposed modification quite significantly (in the most challenging cases, it more than doubles accuracy).


\begin{table}[h]
\centering
\caption{\label{tab:sorting}The sorting experiment: out-of-sample sorting accuracy for various problem sizes and processing steps, with or without glimpses. All the reported accuracies are shown after reaching 10000 training
iterations, at which point all models had converged but none overfitted. Higher is better.}
\begin{tabular}{|l||ll||ll|ll|ll|ll|}
\hline
Length $N$ & \multicolumn{2}{c||}{Ptr-Net}  & \multicolumn{2}{c|}{$P=0$ step} & \multicolumn{2}{c|}{$P=1$ step} & \multicolumn{2}{c|}{$P=5$ steps} & \multicolumn{2}{c|}{$P=10$ steps}\\
\hline
glimpses & \multicolumn{1}{c}{0} & \multicolumn{1}{c||}{1}  & 0 & 1 & 0 & 1 & 0 & 1 & 0 & 1\\
\hline\hline
$N=5$ &  81\% &  90\% &  65\% &  84\% &  84\% &  92\% &  88\% &  94\% &  90\% &  94\%\\
\hline
$N=10$ &  8\% &  28\% &  7\% &  30\% &  14\% &  44\% &  17\% &  57\% &  19\% &  50\%\\
\hline
$N=15$ &  0\% &  4\% &  1\% &  2\% &  0\% &  5\% &  2\% &  4\% &  0\% &  10\%\\
\hline
\end{tabular}
\end{table}

\section{Output Sets}
\label{sec:output_sets}

So far, we have considered the problem of encoding input sets;
let us now turn our attention to output representations.
The chain rule which describes joint probabilities over sets of random variables $Y$ is, perhaps, the simplest decomposition of the joint probability which does not incur arbitrary restrictions (such as conditional independence). Thus, as long as a powerful model that is trainable exists (which can cope with long range correlations), any order should work without any prior order information of the underlying problem that generated $Y$. Despite this, and even when a very powerful model (in terms of modeling power, and resilience to vanishing long term gradients) like the LSTM is employed, output ordering still plays a key role in successfully training models.

In the next subsection, we describe how the order in which we apply the chain rule affects the performance on various tasks.

\subsection{Output Order Matters}\label{sec:output_order}

Let $Y$ be a set (or a sequence). In this section, we will study the effect that ordering has on the performance of seq2seq models on several tasks. Namely, we will consider arbitrary (and non-arbitrary) orders over the variables in $Y$, and model the conditional probability distribution $P(Y|X)$ following that order for all training examples. As we will see, {\em order matters} (even when considering that the formulation through the chain rule works regardless of the ordering of $Y$, at least in principle).

\subsubsection{Language Modeling}\label{sec:lm}

For this experiment, we use the PennTree Bank, which is a standard language modeling benchmark. This dataset is quite small for language modeling standards, so most models are data starved. We trained medium sized LSTMs with large amounts of regularization (see medium model from~\cite{zaremba_regularization}) to estimate probabilities over sequences of words. We consider three version of the dataset with three orderings: natural, reverse, and a fixed, 3-word reversal:

Natural: ``This is a sentence ."\\
Reverse: ``. sentence a is This"\\
3-word: ``a is This $<$pad$>$ . sentence"

Note that the 3-word reversal destroys the underlying structure of the sentence, and makes modeling the joint probability much more difficult since many higher order n-grams are scrambled. For each ordering we trained a different model. The results for both natural and reverse matched each other at 86 perplexity on the development set (using the same setup as~\cite{zaremba_regularization}). Surprisingly, the 3-word reversal degraded only 10 perplexity points, still achieving an impressive result in this corpus at 96 perplexity. We note, however, that training perplexities were also 10 points higher, which indicates that the model had trouble handling the awkward ordering. Thus, even when considering that the chain rule still properly models the joint probability, some degradation was observed when a confounding ordering was chosen.

\subsubsection{Parsing}

The task of constituency parsing consists in producing a parse tree given a sentence.
The model proposed by~\cite{grammar} is a sentence encoder LSTM followed by a decoder LSTM trained
to generate a depth first traversal encoding of the parse tree, using an attention mechanism.
This approach matched state-of-the-art results on this task.

\begin{figure}[h!]
\centering
\includegraphics[width=0.9\linewidth]{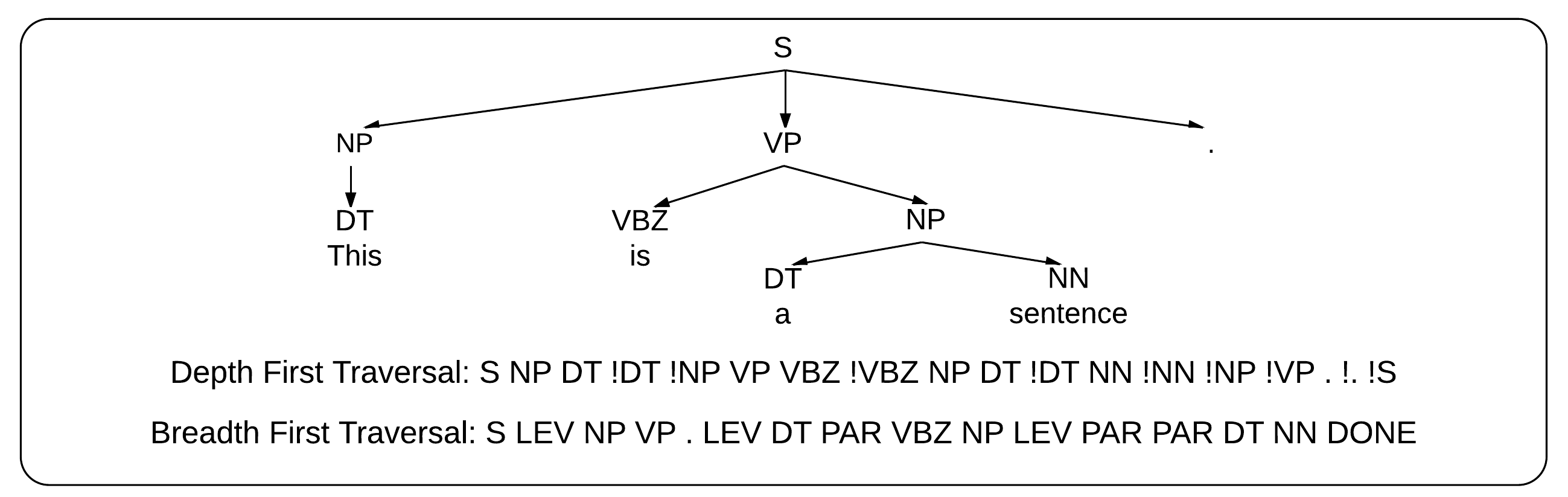}
\caption{\label{fig:dftbft}Depth first and breadth first linearizations of a parse tree which shows our different setups for output ordering in the parsing task.}
\end{figure}

Even though it seemed more sensible, depth first traversal is only one of the many ways one can uniquely encode a tree onto a sequence. We thus tried to train a small model using depth first traversal (which matches the baseline of~\cite{grammar}) and another one using breadth first traversal (note that these orderings are input dependent). See Figure~\ref{fig:dftbft} for an example on how the tree linearizes under both traversal schemes. The model trained to produce depth first traversal linearized trees obtained 89.5\% F1 score (as reported by~\cite{grammar}), whereas the one producing breadth first traversal trees had a much lower F1 score at 81.5\%,\footnote{In fact, in many cases the decoder failed to produce a valid tree, so the real F1 score is likely lower.} showing again the importance of picking the right output ordering.

\subsubsection{Combinatorial Problems}

Unlike in the previous two examples, a problem that more commonly comes up as we try to represent non-sequential data (like tours, triangulations, etc., discussed by~\citep{pointernets}), is the fact that there may exist a large equivalence class of solutions. 

Take, as an example, outputting the indices for the sorted inputs of a set of random numbers, $X$. Indeed, this is a deterministic function. We can choose to output these indices in some order (e.g., increasing, decreasing, or using any arbitrary fixed permutation), or treat them as a set (a tuple of argsort indices with corresponding ranking). As a result, there are $n!$ possible outputs for a given $X$, all of which are perfectly valid. If our training set is generated with any of these permutations picked uniformly at random, our mapping (when perfectly trained) will have to place equal probability on $n!$ output configurations for the same input $X$. Thus, this formulation is much less statistically efficient.

In previous work~\citep{pointernets}, it was found that restricting as much as possible the equivalence class for the outputs was always better. For instance, to output a tour (i.e. a sequence of cities one has to visit for the traveling salesman problem), we started from the lower indexed city (i.e., the first city that we input), and followed a counter-clockwise ordering. Similarly, to output a set of triangles (which triangulate the set of input points), we sorted them in lexicographical order and moved left to right. In all cases, improvements of 5\% absolute accuracy or more were observed.

Failing to restrict the output equivalence class generally implies much slower convergence (and, thus, requires much more training data). For instance, for sorting, if considering the outputs as sets which we output in any of the possible $n!$ orderings, convergence for $n$ as small as 5 never reached the same performance.

\subsubsection{Graphical Models}

Let us consider the joint probability of
a set of $T$ random variables $P(y_1, y_2, \ldots, y_T)$. Having no prior
on how these random variables interact with each other, one way to model
their joint probability is to use the chain rule as follows:
\begin{equation}
P(y_1, y_2, \ldots, y_T) = \prod_{t=1}^T P(y_t | y_1, y_2, \ldots, y_{t-1}) \label{eq:chainrule}
\end{equation}
and model this using an RNN, similar to RNN language models.

While for sentences the natural order of words gives a good clue
of how to order the random variables in the model, for other kind of data
it might be harder to decide on it. Furthermore, in theory, the order should
not matter, because of Bayes rule which lets us reorder all the conditional
probabilities as needed. In practice however, it might be that one order
is easier to model than another, as we have shown in this paper.

The purpose of this experiment is to demonstrate this using a controlled
toy experiment. We generated {\em star-like} graphical models over random
variables where one variable (the head) follows an unconditional distribution,
while the others follow a conditional distribution based on the value of the head
variable. We expect that it should be easier to model the
joint distribution by choosing any ordering which starts with the head variable.
We created several artificial datasets by varying 
the number of random variables to model (between 10 and 50, each of which
was a multinomial over 10 symbols),
the training set size (between 200 and 20000 training samples), and
the {\em randomness} of the marginal distributions, or how deterministic,
or peaky, they were.
For each problem, we trained two LSTMs for 10,000 mini-batch iterations to
model the joint probability, one where the head random variable was shown
first, and one where it was shown last.

The results were as follows:
\begin{itemize}
\item when the training set size is large enough (20000), the LSTM is able to
learn the joint probability in whichever order;
\item when the marginal distributions are very peaky (and thus almost
deterministic), the LSTM is also able
to learn the joint probability independently of the order;
\item in all other cases (small training set size, small or large number
of random variables, and some amount of randomness in the marginal distributions),
it was always easier to learn an LSTM with the {\em optimal} order of random
variables than any other order.
\end{itemize}

\subsection{Finding Optimal Orderings while Training}\label{sec:outsort}

Recall the model we proposed for dealing with input sets: given an embedding for each of the inputs, we have a generic module that is able to process its inputs in any order. This yields an embedding satisfying the key property of being invariant to reorderings, whilst being generic in the kinds of computations to do over the input set.

Unfortunately, placing a joint probability over a set of random variables $y_1,\ldots y_n$ when the structure of the joint probability function is unknown is a hard problem. Fortunately, and thanks to recurrent neural networks, we can apply the chain rule which decomposes this joint probability sequentially (see eq.~\ref{eq:chainrule}) without independence assumptions. In this work, we focus on using the chain rule, discarding more naive decompositions that have strong and unrealistic assumptions (e.g., conditional independence).

An obvious drawback of the chain rule which violates the argument of treating $y_1,\ldots y_n$ as a set is that we condition these random variables in a particular order. Even though, in principle, the order should not matter, in the previous section we have shown that this is indeed not the case, and that certain orderings are better than others in a variety of tasks -- most likely due to the parameterization of the joint probability (using an LSTM), and the non-convex nature of the optimization problem.

Our proposed solution to deal with the aforementioned drawback is extremely simple: as we train, we let the model decide which is the best ordering in which it will apply the chain rule. More formally, assume there exists an ordering which maximally simplifies the task, $\pi(X)$ (where $X$ is the input sequence or set, which can be empty). We would like to train the model as $p(Y_{\pi(X)}|X)$. The number of possible orderings is large -- $n!$ where $n$ is the length of the output, and the best order is unknown a priori.

Since $n!$ can be very large, we could attempt to do (inexact) search as we train the model. Instead of maximizing the log probability of $p(Y|X)$ for each training example pair, we also maximize over orderings as follows:

\begin{equation}
\theta^\star = \arg\max_{\theta} \sum_i \max_{\pi(X_i)} \log p(Y_{\pi(X_i)} | X_i; \theta)\label{eqn:output}
\end{equation}
where $\max_{\pi(X_i)}$ is computed either naively, or with an inexact search. Note that Equation~(\ref{eqn:output}) may not strictly improve the regular maximum likelihood framework due to non-convexity, but we found this to not be an issue in practice.

Besides not being scalable, we found that, if done naively and picking the max over ordering as we train, the model would pick a random ordering (as a function of the initial parameters), and would get stuck on it permanently (since it would reinforce it through learning). We added two ways to explore the space of all orderings as follows:

\begin{itemize}
\item We pretrain the model with a uniform prior over $\pi(X)$ for 1000 steps, which amounts
to replacing the $\max_{\pi(X_i)}$ in eq.~(\ref{eqn:output}) by a $\sum_{\pi(X_i)}$.
\item We then pick an ordering by sampling $\pi(X)$ according to a distribution proportional to $p(Y_{\pi(X)}|X)$. This costs $O(1)$ model evaluations (vs. naive search which would be $O(n!)$).
\end{itemize}

Crucially, sampling $p(Y_{\pi(X)}|X)$ can be done very efficiently as we can use ancestral sampling (left-to-right) which requires to evaluate $p(.)$ only once instead of $n!$.

\subsubsection{5-gram Modeling}

In our initial attempt to solve~(\ref{eqn:output}), we considered a simplified version of the language modeling task described in Section~\ref{sec:lm}. The simplified task consists of modeling the joint probability of 5-grams without any further context (i.e., there is no input $X$). This choice allowed us to have a small enough $n$ as initially we were trying to exactly find the best ordering out of the $n!$ possible ones. Thus, we disregarded possible effects of inexact search, and focused on the essential of the training dynamics where the model being optimized picks the best ordering $\pi$ which maximizes $p(Y_{\pi})$ under its current parameters, and reinforces that ordering by applying updates on the gradient of $\log p(Y_{\pi})$ w.r.t. the parameters. Eventually, as noted in Section~\ref{sec:outsort}, we found sampling to be superior in terms of convergence, whilst simplifying the complexity from $O(n!)$ down to $O(1)$, and is the preferred solution which we used in the rest of this section.

To test this framework, we converted 5-grams (i.e., sequences of words) to a set in the following way:

5-gram (sequence): $y_1$=This, $y_2$=is, $y_3$=a, $y_4$=five, $y_5$=gram \\
5-gram (set): $y_1$=(This,1), $y_2$=(is,2), $y_3$=(a,3), $y_4$=(five,4), $y_5$=(gram,5)

Note that adding the original position alongside the words makes $Y$ a set. Thus, we can shuffle $Y$ in arbitrarily without losing the original structure of the sequence. The first experiment, which reinforces our result in Section~\ref{sec:lm}, tests the hypothesis once again that order matters. Training a model which follows the natural order (i.e., produces (This,1), followed by (is,2) conditioned on (This,1), etc.), achieves a validation perplexity of 225.\footnote{This is much worse than the results reported in Section~\ref{sec:lm} since modeling 5-grams without context is much harder than standard language modeling.} If, instead of picking $(1,2,3,4,5)$, we use $(5,1,3,4,2)$, perplexity drops to 280.

We then test optimization of eq.~(\ref{eqn:output}) in two setups:

Easy: The training set contains examples from $(1,2,3,4,5)$ and $(5,1,3,4,2)$, uniformly sampled.\\
Hard: The training set contains examples from the $5!$ possible orderings, uniformly sampled.

Our results are shown in Table~\ref{tab:output}. Note that, in the easy case, we restrict the search space over orderings to only 2, where one order is clearly better than the other. We note that, after the pretraining phase, we decide which of the two orderings is better to represent the data under the model being trained. Very quickly, the model settles on the natural $(1,2,3,4,5)$ ordering, yielding a perplexity of 225. In the most difficult case, where any order is possible, the model settles to orders such as $(1,2,3,4,5)$, $(5,4,3,2,1)$, and small variations of them. In all cases, the final perplexity is 225. Thus, the framework we propose is able to find good orderings without any prior knowledge. We plan to not only recover optimal orderings, but find ones that were unknown to us when applying the seq2seq framework naively.

\begin{table}[h]
\centering
\caption{\label{tab:output} Experiments in which the model finds the optimal ordering of a set for the 5-gram language modeling task. Perplexities are reported on the validation set (lower is better).}
\begin{tabular}{|l|l|l|}
\hline
Task & Orders considered & Perplexity \\
\hline
$(1,2,3,4,5)$ & 1 & 225 \\
\hline
$(5,1,3,4,2)$ & 1 & 280 \\
\hline
Easy & 2 & 225 \\
\hline
Hard & $5!$ & 225 \\
\hline
\end{tabular}
\end{table}

\section{Conclusion}

LSTMs have shown to be powerful models to represent variable length sequential data thanks to
their ability to handle reasonably long term dependencies and the use of the chain rule to
efficiently decompose joint distributions. On the other hand, some problems are expressed
in terms of an unordered set of elements, either as input or as outputs; in some other cases,
the data is represented by some structure that needs to be linearized to be fed to the LSTM,
and there might be more than one way to do so. The first goal of this paper was to shed some
light on these problems: indeed, we show that {\em order matters} to obtain the
best performance. We then considered the case of unordered input data, for which we proposed the
Read-Process-and-Write architecture, and the case of unordered output data, for which we proposed an efficient
training algorithm that includes a search over possible orders during training and inference.
We illustrated our proposed approaches for input and output sets through various experiments
such as sorting, graphical models, language modeling, and parsing.

\subsubsection*{Acknowledgments}

We would like to thank Ilya Sutskever, Navdeep Jaitly, Rafal Jozefowicz, Quoc Le, Lukasz Kaiser, Geoffrey Hinton, Jeff Dean, Shane Gu and the Google Brain Team for useful discussions on this topic. We also thank the anonymous reviewers which helped improving our paper.

{\small
\bibliography{set2set}
\bibliographystyle{iclr2016_conference}
}

\end{document}